\documentclass[format=acmsmall, review=false, screen=true]{acmart}

\usepackage{booktabs} 

\usepackage[ruled]{algorithm2e} 

\SetAlFnt{\small}
\SetAlCapFnt{\small}
\SetAlCapNameFnt{\small}
\SetAlCapHSkip{0pt}
\IncMargin{-\parindent}

\setcopyright{acmlicensed}
\usepackage{xcolor}

\begin{document}
\title[Energy Disaggregation and Appliance On/Off Detection]{Deep Learning Based Energy Disaggregation and On/Off Detection of Household Appliances}

\author{Jie Jiang}
\author{Qiuqiang Kong*}\thanks{*Corresponding author}
\author{Mark D. Plumbley}
\author{Nigel Gilbert}
\affiliation{%
  \institution{University of Surrey}
  \city{Guildford}
  \state{Surrey}
  \postcode{GU2 7XH}
  \country{United Kingdom}
}
\email{jie.jiang, q.kong, m.plumbley, n.gilbert@surrey.ac.uk}
\author{Mark Hoogendoorn }
\author{Diederik Roijers}
\affiliation{%
  \institution{VU University Amsterdam}
  \city{Amsterdam}
  \postcode{1081 HV}
  \country{Netherlands}
}
\email{m.hoogendoorn, d.m.roijers@vu.nl}

\begin{abstract}
Energy disaggregation, a.k.a. Non-Intrusive Load Monitoring, aims to separate the energy consumption of individual appliances from the readings of a mains power meter measuring the total energy consumption of, e.g. a whole house. Energy consumption of individual appliances can be useful in many applications, e.g., providing appliance-level feedback to the end users to help them understand their energy consumption and ultimately save energy. Recently, with the availability of large-scale energy consumption datasets, various neural network models such as convolutional neural networks and recurrent neural networks have been investigated to solve the energy disaggregation problem. Neural network models can learn complex patterns from large amounts of data and have been shown to outperform the traditional machine learning methods such as variants of hidden Markov models. However, current neural network methods for energy disaggregation are either computational expensive or are not capable of handling long-term dependencies.
In this paper, we investigate the application of the recently developed WaveNet models for the task of energy disaggregation. Based on a real-world energy dataset collected from 20 households over two years, we show that WaveNet models outperforms the state-of-the-art deep learning methods proposed in the literature for energy disaggregation in terms of both error measures and computational cost. 
On the basis of energy disaggregation, we then investigate the performance of two deep-learning based frameworks for the task of on/off detection which aims at estimating whether an appliance is in operation or not. The first framework obtains the on/off states of an appliance by binarising the predictions of a regression model trained for energy disaggregation, while the second framework obtains the on/off states of an appliance by directly training a binary classifier with binarised energy readings of the appliance serving as the target values.
Based on the same dataset, we show that for the task of on/off detection the second framework, i.e., directly training a binary classifier, achieves better performance in terms of F1 score. 
\end{abstract}

%
%


%
%


\maketitle

\renewcommand{\shortauthors}{J. Jiang et al.}

\section{Introduction}

\begin{figure}[!h]
  \centering
  \centerline{\includegraphics[width=\columnwidth]{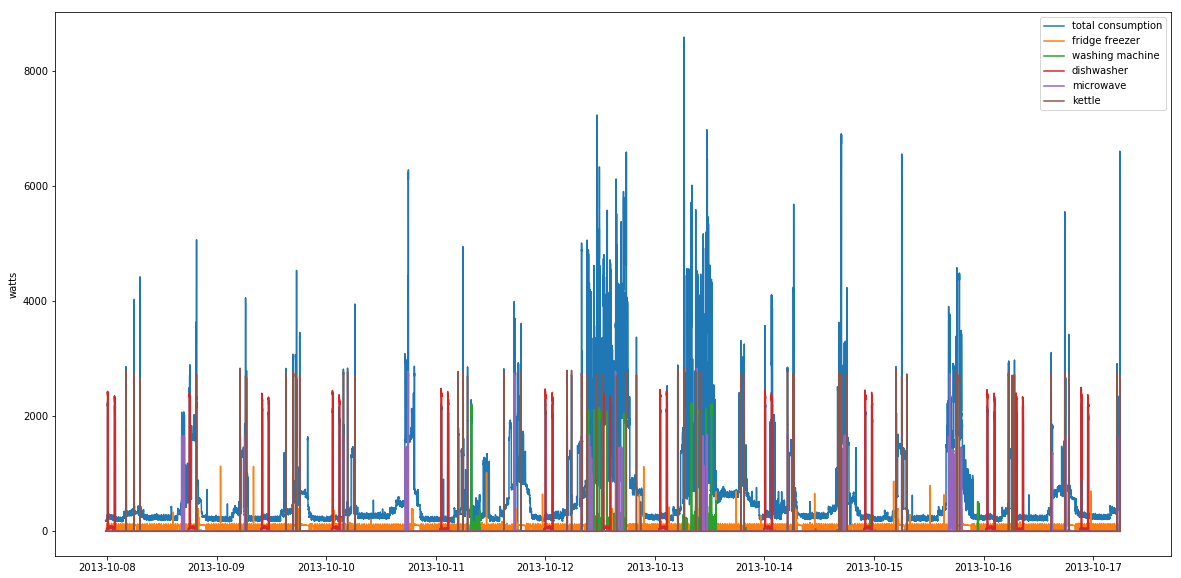}}
  \caption{An example of energy consumption of individual appliances and a whole house}
  \label{fig:energy-dis}
\end{figure}

In recent years, several large-scale household energy consumption datasets are made publicly available, e.g. UK-DALE \cite{UK-DALE} (UK Domestic Appliance-Level Electricity) and REFIT \cite{Murray2017refit} (Personalised Retrofit Decision Support Tools For UK Homes Using Smart Home Technology). These datasets boost the studies on energy disaggregation, also known as Non-Intrusive Load Monitoring (NILM) \cite{Hart1992NILM}. 
Energy disaggregation is a challenging blind source separation problem that aims to separate the energy consumption of individual appliances from the readings of the aggregate meter measuring the total consumption of multiple appliances for example in a house. 
Figure \ref{fig:energy-dis} gives an example of how the energy consumption of a whole house changes along with that of the individual appliances. This problem is difficult due to a number of uncertainties such as the existence of background noise, the lack of knowledge on the numbers of different appliances and their true energy consumption patterns in a given household, replacements of old appliances, and overlapped operations of multiple appliances with similar energy consumption patterns. 

Energy disaggregation finds its usefulness in many applications. 
For example, disaggregated data could be used by feedback systems to provide pertinent information about energy usage and educate consumers at opportune times \cite{Froehlich2009}, which in turn helps the consumers better control their consumption and ultimately save energy \cite{Fischer2008}. Disaggregated data may also help identify malfunctioning equipments or inefficient settings \cite{Froehlich2010}. For policy makers, knowing the amount of energy each category of appliances consumes is critical to the development and evaluation of energy efficiency policies \cite{Sidler1999, Sidler:2003}.
Disaggregated data may also provide valuable information to facilitate power system planning, load forecasting, new types of billing procedures, and the ability to pinpoint the origins of certain customer complaints \cite{Froehlich2010}. 
Another application is to help researchers understand the occurrences of home activities which nowadays are heavily related with the usage of different types of appliances \cite{DSP-2018}. 

In the literature, a lot of research has been done on applying machine learning methods to the problem of energy disaggregation. Among the popular approaches, different variants of HMMs (Hidden Markov Models) such as FHMMs (Factorial HMMs) have attracted a lot of attention \cite{Kolter2012FHMM,Zhong2014NIPS,Shaloudegi2016FHMM}. With the availability of large-scale open datasets such as UK-DALE and REFIT \cite{UK-DALE,Murray2017refit}, there is a flourish in applying deep neural networks (DNNs) to the problem of energy disaggregation. For example, in \cite{Kelly:2015} and \cite{zhang2018sequence}, the authors investigated the application of convolutional neural networks (CNNs), recurrent neural networks (RNNs) and Autoencoders. However, there are several problems with the conventional DNN models. The computation complexity of the conventional CNNs is getting substantially high when the input sequences are long. In the case of RNNs, the values of the hidden units have to be calculated in a sequential order and thus does not scale well. Recently, a neural network called WaveNet \cite{van2016wavenet} was proposed for long sequence audio processing. WaveNet is a variant of the CNN architecture with dilated convolutional layers which make it easier to be trained with long sequences compared to the conventional CNNs. With skip connections over all the convolutional layers, it can learn multi-scale hierarchical representations. WaveNet has been proven to work well for tasks such as speech synthesis \cite{van2016wavenet} and speech denoising \cite{rethage2017wavenet}. It is efficient because it has fewer parameters than a CNN does. WaveNet is also easy to parallelise compared to RNNs. For the task of energy disaggregation, some appliances may have long-term dependencies in their energy consumption patterns and these patterns may exist at different scales. Therefore, the task of energy disaggregation may benefit from WaveNet's capability of modeling long sequences and learning multi-scale hierarchical representations.

To evaluate the performance of WaveNet models for the task of energy disaggregation, we carried out a set of experiments using the public dataset REFIT \cite{Murray2017refit}, and compared the disaggregation results of WaveNet models against the five-layer CNN model proposed in \cite{zhang2018sequence} and a three-layer RNN model. We showed that  WaveNet models outperform the other two methods in terms of both error measures and computation cost. We also investigated the influence of the length of input sequences on the disaggregation performance as well as on the computation cost.

While the problem of energy disaggregation focuses on estimating the exact amount of energy being consumed by an appliance, there is another interesting task called on/off detection which tries to predict whether individual appliances are in operation or not. Compared to energy disaggregation, on/off detection provides a perspective of a coarser granularity on the usage status of individual appliances and finds its usefulness in understanding the occurrences of home activities that heavily depend upon the assistance of household appliances such as kettle \cite{Alcala:2015}, washing machine and microwave \cite{DSP-2018}. Such dependencies have been proven, for example, to help with activity monitoring and health management \cite{Alcala:2015}.
In this paper, we investigate two learning frameworks for the task of on/off detection. The first one, called regression based learning framework, first trains a model for energy disaggregation using the aggregate energy readings as inputs and the appliance readings as the target values, and then derives the on/off state sequence of the appliance by binarising the predictions of the disaggregation model according to the on-power threshold of the appliance. The second one, called classification based learning framework, directly trains a binary classifier with the appliance on/off states.

To evaluate the two learning frameworks for the task of on/off detection, we respectively trained a group of WaveNet models following the two learning frameworks with the REFIT dataset. We showed that for the task of on/off detection the classification based learning framework outperforms the regression based learning framework in terms of F1 score.

The contributions of this paper are as follows:
\begin{itemize}
\item We propose to tackle the problem of energy diaggregation with  WaveNet models which are capable of modeling long sequences more efficiently compared to the conventional CNNs and RNNs, and we show that WaveNet models achieve the state-of-the-art performance based on a set of experiments with a public dataset. 
\item We carry out an analysis on how the receptive field size and the target field size would affect the disaggregation performance of the three deep neural networks, i.e., WaveNets, CNNs and RNNs.
\item We compare a regression based learning framework with a classification based learning framework for the task of on/off detection and show empirically that the latter outperforms the former that utilises the outputs from energy disaggregation. 
\item The evaluation is performed using the public dataset REFIT collected from 20 households. We give a detailed description of how the raw data was preprocessed and used for model training and release the source code\footnote{https://github.com/jiejiang-jojo/fast-seq2point} to facilitate the reproducibility of our work. 
\end{itemize}

The rest of the paper is organized as follows: Section \ref{relatedwork} discusses the related work. Section \ref{problem} gives a formal description of the energy disaggregation problem and the on/off detection problem. Section \ref{methods} presents three learning paradigms for model training, introduces three neural network models, and describes how a model is trained respectively for the task of energy disaggregation and on/off detection. Section \ref{experiments} illustrates the experiment preparation and analyses the experiment results. Finally, in Section \ref{conclusion}, we conclude the paper with possibilities of future work.

\section{Related Work}\label{relatedwork}
In the literature, a lot of research has been done on applying machine learning methods to the problem of energy disaggregation. Among the popular approaches, different variants of hidden Markov models (HMMs) have attracted much attention (e.g. \cite{Kim2011, Kolter2012FHMM, Parson2012Prior, Zhong2014NIPS, Shaloudegi2016FHMM}). 
Recently, with the availability of large open datasets such as UK-DALE and REFIT \cite{UK-DALE,Murray2017refit} and the superior performance of deep neural networks (DNNs) in many research areas such as computer vision \cite{krizhevsky2012imagenet} and audio processing \cite{rethage2017wavenet}, there has been a flourish on applying DNNs for the problem of energy disaggregation. For example, Kelly and Knottenbelt \cite{Kelly:2015} compared the disaggregation performance of the traditional machine learning methods (e.g. FHMMs) with the deep learning methods such as Autoencoders and Long Short-term Memory (LSTM) networks and the results show that the deep learning methods outperform the traditional methods. Mauch and Yang \cite{Mauch15} also advocated the application of LSTM for the problem of energy disaggregation. Chen et al. \cite{Chen2018} proposed a convolutional sequence to sequence model in which gated linear unit convolutional layers were used to extract information from the sequences of aggregate electricity consumption and residual blocks were used to refine the output of the neural network. Later, Zhang et al. \cite{zhang2018sequence} proposed to use a sequence-to-point paradigm to train a CNN for energy disaggregation which outperforms the sequence-to-sequence learning approach used in \cite{Kelly:2015}. 
There are also works using a combination of DNNs. For example, by combining CNNs with variational autoencoders, Sirojan et al. \cite{Sirojan18} showed that their approach outperforms the one presented in \cite{zhang2018sequence}. Shin et al. \cite{Shin2019} proposed a subtask gated network that combines the main regression network with an on/off classification subtask network. 
Targeting real-time applications, Harell et al. \cite{Harell2019} proposed a causal 1-D CNN based on the WaveNet model proposed in \cite{van2016wavenet}. This work is similar to ours as it also adapts WaveNet for the problem of energy disaggregation but our work differs from this work as we use a non-causal version of WaveNet proposed in \cite{rethage2017wavenet}, i.e., the same amount of samples in the past as well as in the future are used to train the model and inform the prediction. Another difference is that we employed the concept of target field as proposed in \cite{rethage2017wavenet} such that the computation of the neighbouring samples can be shared, which speeds up the model training and inference. In addition, we carried out an extensive study on how the receptive field size and the target field size influence the model performance. 
Moreover, the baseline used in \cite{Harell2019} is a variant of HMM, i.e. sparse super-state HMM, while we compared our work with the state-of-the-art DNN based approaches.

\section{Problem Statement}\label{problem}

\subsection{Energy Disaggregation}\label{energy_dis}
Energy disaggregation aims to estimate the energy usage of individual appliances based on the readings of the mains power meter that measures the total energy consumption of, for example, a whole house. Formally, suppose we have a sequence of readings from a house-level meter denoted as $X=(x_1, x_2, \ldots, x_T)$ where $ T $ is the length of the sequence. The problem of energy disaggregation is to disaggregate $ X $ into the energy consumption sequence of individual appliances denoted as $ Y^i=(y_{1}^{i}, y_{2}^{i}, \dots, y_{T}^{i})$, $y^i_t \in \mathbb{R}_{\geq 0} $ where $ \mathbb{R}_{\geq 0} = [0, +\infty) $, $ I $ is the number of known appliances, $i \in \{1, \ldots, I\}$ is the appliance index and $ t \in \{1, \ldots, T\}$ is the index of samples in time domain. In addition, we denote the readings from unknown appliances and background noise as $ U = (u_{1}, ..., u_{T}) $. At any time $t$, $x_t$ is assumed to be the summation of the readings from all the known appliances and unknown appliances with background noise:

\begin{equation} \label{eq:mae}
x_{t} = \sum_{i=1}^{I} y^{i}_{t}+u_{t}
\end{equation}

\noindent where the residual term $ u_{t} $ indicates the energy consumption of unknown appliances and background noise at time $t$. 
The aim of energy disaggregation is to design a model to separate the energy consumption of the individual appliances $ Y^{i}, i \in \{1, \ldots, I\} $ from the aggregate readings $ X $. That is, we are looking for a set of disaggregation mappings: 

\begin{equation} \label{eq:mapping_disaggregation}
f^{i}: X \mapsto Y^{i}.
\end{equation}
where each mapping $ f^{i} $ maps from an aggregate reading sequence $ X $ to the energy consumption sequence of an appliance $ Y^{i} $.

\subsection{On/off Detection}\label{status_detection}

On a coarser granularity, most appliances have an on-power threshold which defines the least amount of energy an appliance needs to operate. When the amount of energy an appliance is consuming is below the on-power threshold it is considered to be in an off state otherwise it is considered to be in an on state. For example, a kettle usually needs 2000 watts to be in an on state while a washing machine needs 20 watts. The problem of on/off detection aims to estimate whether an appliance is in an on or off state based on the readings of the total energy consumption. 

Formally, given a sequence of aggregated readings from a house-level meter denoted as $X=(x_1, x_2, \ldots, x_T)$ where $ T $ is the length of the sequence, 
the aim of on/off detection is to design a model to recognise the on/off state sequence of individual appliances denoted as $ Z^i=(z_{1}^{i}, z_{2}^{i}, \dots, z_{T}^{i})$, $z^i_t \in \{0,1\}$, $i \in \{1, \ldots, I\}, t \in \{1, \ldots, T\}$ from the aggregate readings $ X $ where $ I $ is the number of known appliances, $ i $ is the appliance index, $ t $ is the sample index in time domain, $0$ indicates the off state and $1$ indicates the on state. That is, we are looking for a set of mappings:  

\begin{equation} \label{eq:mapping_onoff}
f^{i}: X \mapsto Z^{i}.
\end{equation}
where each mapping $ f^{i} $ maps from an aggregate reading sequence $X$ to the on/off state sequence of an appliance $Z^{i}$.

\section{Methods}\label{methods}

\subsection{Learning Paradigms}\label{learning_paradigms}
Usually the aggregated readings $ X $ is a long sequence over days, months, or sometimes years. To efficiently train a model, a commonly used approach in the literature of energy disaggregation is the sliding window approach which splits the long sequence $ X $ into shorter sequences $ \mathbf{x} = (x_{t}, ..., x_{t+L-1}) $ where $ L $ indicates the receptive field size. Instead of learning the mappings in Equation \ref{eq:mapping_disaggregation} and Equation \ref{eq:mapping_onoff} directly, we use sequences $ \mathbf{x} $ as input. The target of an input sequence can be a sequence of the same length which is called sequence-to-sequence learning  \cite{Kelly:2015} or the midpoint of the target sequence which is called sequence-to-point learning  \cite{zhang2018sequence}. 
In this section, we introduce three variants of the sliding window approach.

\subsubsection{Sequence-to-sequence Learning}
Sequence-to-sequence learning \cite{Kelly:2015}, as shown in Figure  \ref{fig:fast_seq_to_point} (a), was proposed to learn a mapping from an input sequence $ \mathbf{x} $ to an output/target sequence $ \mathbf{y} $, where $ \mathbf{x}=(x_{t}, ..., x_{t+L-1}) $ and $ \mathbf{y}=(y_{t}, ..., y_{t+L-1}) $ have the same length. In sequence-to-sequence learning, each element of the output signal is predicted many times and an average of these predictions is used as the final output, which consequently smooths the edges.
However, as pointed in \cite{zhang2018sequence}, it is expected that some of the input sequences will provide a better prediction of a single point than others, particularly those sequences where the point is near the midpoint of the input sequence.

\subsubsection{Sequence-to-point Learning}
Sequence-to-point learning \cite{zhang2018sequence}, as shown in Figure \ref{fig:fast_seq_to_point} (b), aims to solve the problem of sequence-to-sequence learning by finding a mapping from an input sequence $ \mathbf{x}=(x_{t}, ..., x_{t+L-1}) $ to a single target point $ y_{t+ \left \lfloor L/2 \right \rfloor} $ the index of which corresponds to the midpoint of the input sequence, where $ \left \lfloor \cdot \right \rfloor $ floors a value to an integer. One problem of the sequence-to-point learning paradigm is that learning a single point is usually inefficient. 

\subsubsection{Fast Sequence-to-point Learning}
\begin{figure}[!h]
  \centering
  \centerline{\includegraphics[width=0.7\columnwidth]{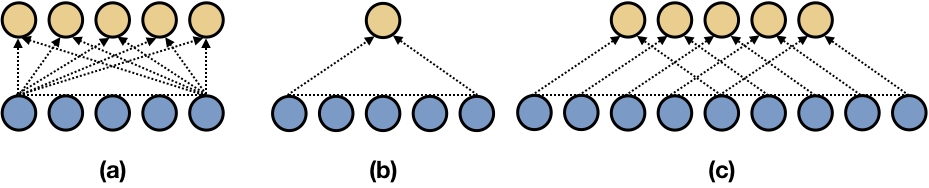}}
  \caption{(a) Sequence-to-sequence learning; (b) Sequence-to-point learning; (c) Fast sequence-to-point learning. }
  \label{fig:fast_seq_to_point}
\end{figure}

In this paper, we propose to use a fast sequence-to-point learning paradigm, as shown in Figure \ref{fig:fast_seq_to_point} (c), to speed up the sequence-to-point learning. By introducing a target field \cite{rethage2017wavenet} to replace a single point as output, the computation of a sequence-to-point model can be shared. The input sequence and target sequence are denoted as $ \mathbf{x}=(x_{t}, ..., x_{t+L+r-2}) $ and $ \mathbf{y}=(y_{t+\left \lfloor L/2 \right \rfloor}, ..., y_{t+\left \lfloor L/2 \right \rfloor+r-1}) $ respectively. The length of the input sequence in this case is $(L+r-1)$ where $L$ indicates the size of the receptive field, and $r$ indicates the size of the target field, i.e. the length of the target sequence. When $r=1$, fast sequence-to-point learning degenerates to sequence-to-point learning. 

\subsection{Deep Neural Networks} \label{sec:NueralNetworks}
In this section, we introduce three classes of neural network. The first two, CNNs and RNNs, serving as two baselines are used to benchmark the performance of WaveNets.

\subsubsection{Convolutional neural networks}\label{cnn}
Convolutional neural networks (CNNs) have achieved the state-of-the-art performance in many applications such as computer vision \cite{krizhevsky2012imagenet}, speech and audio processing \cite{rethage2017wavenet} and natural language processing \cite{dauphin2016language}. With shared filters to capture local patterns of various signals, the number of parameters of a CNN is fewer than that of a fully connected neural network. 

Time domain CNNs have been applied to energy disaggregation, for example, in \cite{zhang2018sequence}. Similar to the two dimensional CNN for computer vision \cite{krizhevsky2012imagenet}, a time domain CNN consists of several convolutional layers, each of which contains several filters that are used to convolve with the output of the previous convolutional layer. The filters are designed to capture local patterns of a signal. For example, in computer vision, the lower level filters of a CNN may learn edge detectors, while the higher level filters may learn to capture high-level profiles of an image. Similarly, in the case of time domain CNNs for energy disaggregation, lower level filters may capture the short-term energy usage patterns of different appliances such as a single activation, while higher level filters may capture the long-term patterns such as a complete operating cycle. 

The time domain convolutional operation can be described as follows:

\begin{equation} \label{eq:cnn}
v[k^{out}, t] = \sum_{k^{in}} \sum_{\tau=1}^{m} u[k^{in}, t - \tau] \cdot h[k^{out}, k^{in}, \tau] 
\end{equation}

\noindent where $ u $ and $ v $ denote the input and output feature maps of a convolutional layer, and $ k^{in} $ and $ k^{out} $ indicate the index of the input and output feature maps. The filters are represented as a three dimensional tensor $ h $ and $ m $ is the filter length in time domain. The first convolutional layer takes a sequence $ \mathbf{x} $ as input. The predicted output is obtained from the last convolutional layer of a CNN. 

With larger receptive fields, long-term dependencies in the energy consumption data can be taken into account. However, the computation complexity of CNNs will increase quadratically along with the size of the receptive field.

\subsubsection{Recurrent neural networks}\label{rnn}

Recurrent neural networks (RNNs) have many successful applications in modeling temporal signals, e.g., audio and speech signal processing \cite{graves2013speech} and natural language processing \cite{chung2014empirical}. Similar to the fully connected neural networks, each input sample $ x_{t} $ is mapped to a hidden unit $ h_{t} $ by a transformation matrix. In addition, there are connections between adjacent hidden units to carry on the information from previous samples. In a non-causal system, a RNN can be bidirectional so as to use information from both history and future. A recurrent layer of a RNN can be described as:

\begin{equation} \label{eq:rnn_layer}
h_{t} = \phi(Wx_{t} + Vh_{t-1} + b)
\end{equation}

\noindent where $ W $, $ V $ and $ b $ respectively represent the transformation matrix between input samples and hidden units, the transformation matrix between adjacent hidden units, and a bias term; $ \phi $ represents a non-linear function. A RNN may consist of several recurrent layers. The backpropagation through time algorithm \cite{werbos1990backpropagation} is used for training a RNN. 

One problem of the conventional RNNs is gradient vanishing/explosion \cite{pascanu2013difficulty}. This is because the depth of a RNN is proportional to the length of the input sequence. When training a RNN, the gradient will accumulate exponentially, which makes the training unstable. To solve the gradient explosion/vanishing problem, LSTM was proposed, which introduces a memory cell with update, forget and output gates to control the information flow \cite{hochreiter1997long}. Later Gated Recurrent Unit (GRU) \cite{chung2014empirical} was proposed to simplify LSTM by reducing the number of parameters. A GRU is described as follows:

\begin{equation} \label{eq:gru}
\begin{matrix}
r_{t} = \sigma(W_{r}x_{t} + U_{r}h_{t-1} + b_{r}) \\ 
z_{t} = \sigma(W_{z}x_{t} + U_{z}h_{t-1} + b_{z}) \\ 
\widetilde{h}_{t} = \phi(Wx_{t} + U(r_{t} \odot h_{t-1}) + b) \\ 
h_{t} = z_{t} \odot h_{t-1} + (1 - z_{t}) \odot \widetilde{h}_{t}.
\end{matrix}
\end{equation}
where $ r_{t} $ indicates the reset gate at time step $t$, $ z_{t} $ indicates the update gate at time step $t$, $\widetilde{h}_{t}$ indicates the candidate new value for the memory cell at time step $t$, $h_{t}$ indicates the final value for the memory cell at time step $t$, $ \sigma $ represents a sigmoid (non-linear) function and $ \phi $ represents a $\tanh$ (non-linear) function. The units that learn to capture short-term dependencies will tend to have reset gates frequently active while the units that learn to capture long-term dependencies will tend to have update gates frequently active. 

\subsubsection{WaveNets}\label{wavenet}

Conventional CNNs cannot scale when the input sequences are getting long as the computation complexity increases quadratically along with the receptive field size. Compared to CNNs, the computation complexity of RNNs increases linearly along with the receptive field size. However, the hidden units of RNNs can only be calculated sequentially because the calculation of each hidden unit depends upon the value of the previous hidden unit. So RNNs cannot handle parallel computations efficiently. Therefore, long sequence modeling has been a computational challenge for both CNNs and RNNs. 

\begin{figure*}[!h]
  \centering
  \centerline{\includegraphics[width=\columnwidth]{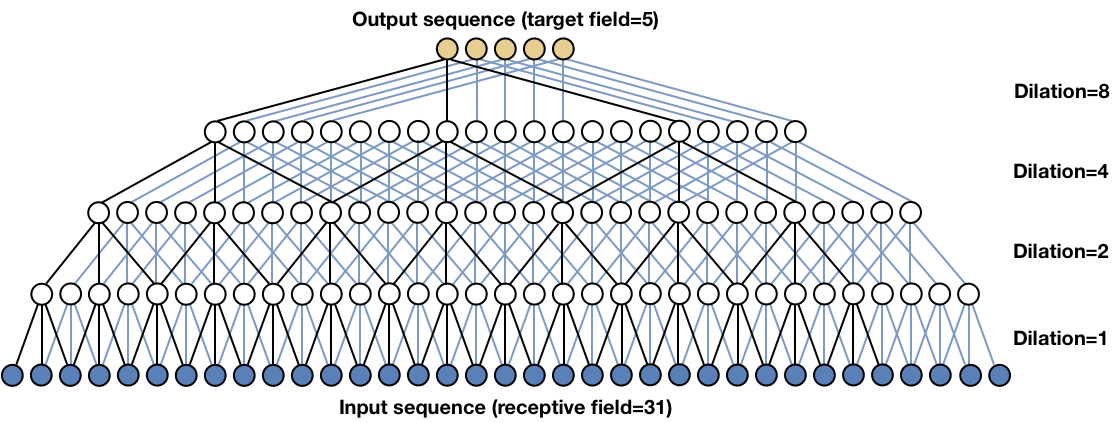}}
  \caption{An example of the WaveNet input-output structure for energy disaggregation. }
  \label{fig:wavenet}
\end{figure*}

\begin{figure}[!h]
  \centering
  \centerline{\includegraphics[width=0.5\columnwidth]{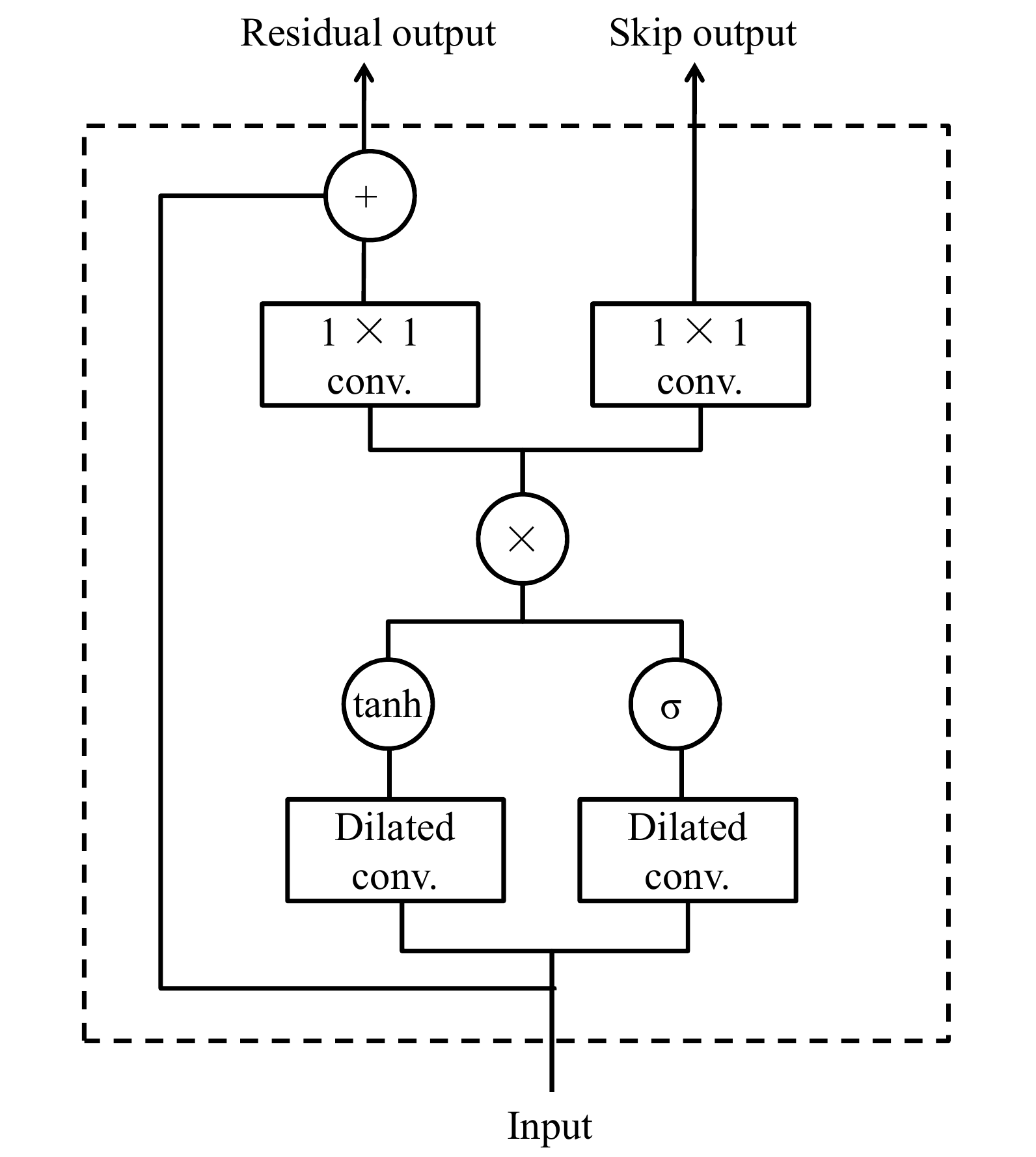}}
  \caption{Residual block of WaveNets. }
  \label{fig:residualblock}
\end{figure}

To solve this problem, WaveNet \cite{van2016wavenet} was proposed for modeling raw audio signals and has been used for modeling time sequences in tasks such as speech denoising \cite{rethage2017wavenet}. WaveNet is an improvement over conventional CNNs, where a ``dilated convolution'' is applied to reduce the size of filters. A dilated convolution is a convolution with holes. That is, the filters are applied over an area larger than its length by skipping input values with a certain step size. Stacked dilated convolutions enable a network to have very large receptive fields with just a few layers. In \cite{van2016wavenet} a filter length of 2 is applied for modeling the causal audio signals. In this paper, following \cite{rethage2017wavenet}, we applied a filter length of 3 to utilise the non-causal information of input sequences. Figure \ref{fig:wavenet} shows through an example the WaveNet input-output structure used in this paper for both tasks of energy disaggregation and on/off detection. Following \cite{van2016wavenet}, the dilated layers are embedded into residual blocks, as shown in Figure \ref{fig:residualblock}. The residual output of each block will be sent to the input of the next one. The skip output of all the residual blocks will be summed and then followed by a $ 3 \times 1 $ convolutional layer, which gives the final output as predictions. 

The following equation characterises the relation between the number of dilated convolutional layers of a WaveNet and its receptive field size:

\begin{equation} \label{eq:seql_layer}
L = (2^s - 1) * (m - 1) + 1
\end{equation}
where $L$ denotes the receptive field size, $s$ denotes the number of dilated convolutional layers, $m$ denotes the filter length applied to each dilated convolutional layer and in this paper we set $m=3$ following \cite{rethage2017wavenet}. 

Since WaveNets do not have recurrent connections \cite{van2016wavenet}, they are typically faster than RNNs, especially when applied to long sequences.

\subsection{Training a Model For Energy Disaggregation}\label{learning_energy_dis}

For the task of energy disaggregation, the training of a fast sequence-to-point model based on CNN/RNN/WaveNet can be implemented with back-propagation. The inputs are sequences of aggregate energy readings $\mathbf{x}$ while the target values are sequences of appliance energy readings $\mathbf{y}$. Assuming an output and the corresponding target value are denoted as $ \mathbf{\hat{y}}=(\hat{y}_{t+\left \lfloor L/2 \right \rfloor}, ..., \hat{y}_{t+\left \lfloor L/2 \right \rfloor+r-1}) $ and $ \mathbf{y}=(y_{t+\left \lfloor L/2 \right \rfloor}, ..., y_{t+\left \lfloor L/2 \right \rfloor+r-1}) $ respectively, the loss can then be calculated using Mean Absolute Error (MAE) which is used as one of the evaluation criteria (see Section \ref{metric} for details):

\begin{equation} \label{eq:loss}
loss(\mathbf{\hat{y}}, \mathbf{y}) = \frac{1}{r} \sum_{\tau=\left \lfloor L/2 \right \rfloor}^{\left \lfloor L/2 \right \rfloor+r-1} \left | \hat{y}_{t+\tau} - y_{t+\tau} \right |. 
\end{equation}
The loss function is calculated on mini-batch data. When the target field size $r$ equals 1, Equation \ref{eq:loss} degenerates to the conventional sequence-to-point model. After obtaining the loss, the gradient can be calculated and used to update the parameters of the model. 

\subsection{Training a Model for On/off Detection}\label{sec:learning_onoff}

\subsubsection{Regression Based Learning Framework}
The regression based learning framework tackles the problem of on/off detection by utilising the outputs from energy disaggregation. In concrete, for a given appliance, it first trains a fast sequence-to-point model based on CNN/RNN/WaveNet for energy disaggregation. Thereafter, given a new sequence of aggregate energy readings, the energy readings of the appliance are predicated using the disaggregation model. Finally, it derives the on/off state sequence of the appliance by binarising the predictions according to the on-power threshold of the appliance. 

\subsubsection{Classification Based Learning Framework}
The classification based learning framework tackles the problem of on/off detection by directly training a binary classifier. In concrete, it first binaries the energy readings of a given appliance according to the on-power threshold of the appliance. Thereafter, it trains a binary classifier using the aggregate energy readings as inputs and the binarsied appliance readings as the target values. Finally, given a new sequence of aggregate energy readings, the on/off state sequence of the appliance is predicated using the trained classifier.

For training a fast sequence-to-point binary classifier for a given appliance, the last layer of a CNN/RNN/WaveNet is a fully connected layer followed by a sigmoid nonlinearity to represent the probability that the appliance is in the on state. Assuming an output and the corresponding target value are denoted as $ \mathbf{\hat{z}}=(\hat{z}_{t+\left \lfloor L/2 \right \rfloor}, ..., \hat{z}_{t+\left \lfloor L/2 \right \rfloor+r-1}) $ and $ \mathbf{z}=(z_{t+\left \lfloor L/2 \right \rfloor}, ..., z_{t+\left \lfloor L/2 \right \rfloor+r-1}) $ respectively, the loss can then be calculated using the binary cross-entropy: 

\begin{equation} \label{binary_crossentropy}
\text{loss}(\mathbf{\hat{z}}, \mathbf{z}) =  - \frac{1}{r} \sum_{\tau=\left \lfloor L/2 \right \rfloor}^{\left \lfloor L/2 \right \rfloor+r-1} (z_{t+\tau} \ \text{ln} \ \hat{z}_{t+\tau} + (1 - z_{t+\tau}) \text{ln}(1 - \hat{z}_{t+\tau})).
\end{equation}
Similarly, the loss function is calculated on mini-batch data. When the target filed size $r$ equals 1, Equation \ref{binary_crossentropy} degenerates to the conventional sequence-to-point model. After obtaining the loss, the gradient can be calculated and used to update the parameters of the model.

\section{Experiments}\label{experiments}

\subsection{Dataset}

The dataset used in this paper is REFIT \cite{Murray2017refit} which is a collection of energy consumption data from 20 households in the UK. The readings were recorded around every 8 seconds and covers a period of over 2 years. The dataset contains both house-level energy usage (aggregate readings) and appliance-level energy usage (appliance readings) of more than 10 types of appliances. In this paper we focus on the disaggregation of four types of appliances: kettle, microwave, dish washer and washing machine which are used by most of the households. 

\subsection{Data Preprocessing}

Firstly, we resampled the data with an interval of 10 seconds to mitigate the fluctuations of time intervals between the original readings, which resulted in 93,976,578 data points. Secondly, following \cite{Kelly:2015}, we filled the gaps in the data shorter than 3 minutes by forward-filling assuming that the gaps are caused by RF issues and filled the gaps longer than 3 minutes with zeros assuming that the gaps are caused by the appliance being switched off. Thirdly, for each type of appliance and the aggregate, we normalised the data by subtracting the mean values and dividing by the corresponding standard deviations. 

Thereafter, for each household, we extracted all the possible segments of length $(L+r-1)$ from the aggregate readings by a sliding window of step-size $r$, where $L$ indicates the size of the receptive field and $r$ indicates the size of the target field. These segments of aggregate readings are used as inputs for training and testing.  For each of the aggregate segments, we obtained the corresponding target sequence by extracting a segment of consecutive appliance readings of length $r$ such that the center of the two segments are aligned. Moreover, we remove any input sequence and its corresponding target sequence where the target sequence contains an appliance reading that is larger than the corresponding aggregate reading in the input sequence.    
Since not every household has all the four appliances, we used the data from the last four households for testing and the data from the rest of the households for training, as shown in Table \ref{tab:household}. 

\begin{table*}[!h]
  \centering
  \caption{Households used for training and testing per appliance.}
  \label{tab:household}
  \begin{tabular}{c|cc}
    \toprule
     Appliance & Training household ID & Test household ID \\
    \midrule
    Kettle              & [2, 3, 4, 5, 6, 7, 8, 9, 11, 12, 13]  & [17, 19, 20, 21] \\
    Microwave           & [2, 3, 4, 5, 6, 8, 9, 10, 11, 12, 15] & [17, 18, 19, 20] \\
    Dishwasher          & [1, 2, 3, 5, 6, 7, 9, 10, 11, 13, 15] & [16, 18, 20, 21] \\
    Washing M.     & [1, 2, 3, 5, 6, 7, 8, 9, 10, 11, 13, 15, 16, 17] & [18, 19, 20, 21] \\
  \bottomrule
\end{tabular}
\end{table*}

As for on/off detection, instead of normalising the appliance readings, we obtain the output data by binarising the appliance readings using the on power thresholds shown in Table \ref{tab:data} in accordance with the previous studies \cite{zhang2018sequence} and \cite{Kelly:2015}.  

\begin{table*}[!h]
  \centering
  \caption{On power threshold for each appliance in watts.}
  \label{tab:data}
  \begin{tabular}{c|ccccc}
    \toprule
     & Kettle & Microwave & Dishwasher & Washing M.\\
    \midrule
    On power threshold      & 2000 & 200 & 10 & 20\\
  \bottomrule
\end{tabular}
\end{table*}

\subsection{Evaluation Metrics}\label{metric}
For the task of energy disaggregation, we used two metrics for evaluation in this paper, i.e. Mean Absolute Error (MAE) and normalised Signal Aggregate Error (SAE). MAE is a measurement of errors that averages over the differences between all the predictions with respect to the real consumptions, which is less sensitive with outliers. SAE is a measurement of errors that sums all the differences between the predictions and the real consumptions over a period of time, e.g. a day, a week, a month etc. In our case, the evaluation is over the whole time period of the testing households' data collection. The formal definitions of MAE and SAE are as follows:

  \begin{equation} \label{mae}
    MAE = \frac{1}{T}\sum_{t=1}^{T} |\hat{y}_t-y_t|
  \end{equation}
where $\hat{y}_t$ indicates the prediction of an appliance's energy usage at time $t$ and $y_t$ indicates the corresponding ground truth.
  
  \begin{equation} \label{sae}
    SAE = \frac{|\hat{r}-r|}{r}
  \end{equation}
where $\hat{r}=\sum_t{\hat{y}_t}$ and $r=\sum_t{y_t}$ respectively indicate the predicated energy consumption of an appliance over a certain time period and the corresponding ground truth.

For the task of on/off detection, we used F1 score to evaluate the performance of different models as the dataset is extremely imbalanced. For example, kettle is on only for about 1\% of the time. F1 score 
\cite{Jeni:2013} can be interpreted as a harmonic average of the precision and recall:

\begin{equation}\label{f1}
F1 = 2 \times \frac{
  precision \times recall
  }{
  precision + recall
  }.
\end{equation}
where precision is the fraction of true positive instances among the predicted positive instances, while recall is the fraction of true positive instances over the total number of positive instances.

In the rest of this paper, when evaluating a model using the metrics above, we remove from the test set all the pairs of aggregate reading and appliance reading in which the aggregate reading is less than the individual appliance reading or the aggregate reading is zero.  

\subsection{Experimental Results For Energy Disaggregation}
\subsubsection{Experiment Setup}
For energy disaggregation, we trained three groups of neural network models. The first group is based on our implementation of the 5-layer CNN proposed in \cite{zhang2018sequence}. The second group is based on a 3-layer bidirectional RNN with GRUs. The third group is based on the WaveNet as shown in Section \ref{wavenet}. We use the Adam optimizer \cite{Kingma:2014} with a learning rate of 0.001 to minimise the loss as shown in Equation \ref{eq:loss}. These hyper-parameters are chosen experimentally. 

For each group of models, we explored the influence of two parameters. The first parameter is the receptive field size $L$. In this paper, we used input sequences with a range of receptive field sizes \textit{15}, \textit{31}, \textit{63}, \textit{127}, \textit{255}, \textit{511}, \textit{1023}, \textit{2047}, which corresponds to numbers of layers \textit{3}, \textit{4}, \textit{5}, \textit{6}, \textit{7}, \textit{8}, \textit{9}, \textit{10} in WaveNet models. Note that the receptive field size of \textit{1023} and \textit{2047} were not applied to the CNN and RNN models for the sake of computation cost in training. The second parameter is the target field size $r$ for which we experimented with four different values \textit{1}, \textit{10}, \textit{100} and \textit{1000}. We used a mini-batch size of \textit{128} for training all the models. 

For most appliances, the duration that an appliance is being used is much smaller than it is not, i.e., the readings are extremely imbalanced between those representing the appliance is in use and those representing it is not. For example, the readings that are less than 10 watts is around 99\% for kettle. In such cases, a model that always predicts a very small value, e.g. zero, may perform well in terms of MAE. Therefore, we employ a naive baseline model, i.e., always predicting zero (always-zero). The metric SAE focuses on the total energy consumption over a period of time, which makes the mean value of an appliance's energy consumption a promising prediction. To this end, we employ another naive baseline model, i.e., always predicting the mean value (always-mean). 

\subsubsection{Result Analysis}
Table \ref{tab:result} shows the best MAE together with the corresponding SAE achieved by the models within each group with a fixed target field size of 100. We can see that the WaveNet model achieves the best MAE over all the four appliances. In particular for \textit{dishwasher} and \textit{washing machine}, the WaveNet model reduces the MAE by 51\% and 38\% comparing to the CNN model while by 32\% and 14\% comparing to the RNN model. As for \textit{kettle} and \textit{microwave}, the WaveNet model and the RNN model obtain similar MAEs. In the case of SAE, the WaveNet model and the RNN model achieve similar results except for the case of \textit{dishwasher} where the WaveNet model has an improvement of 49\%. Overall, the WaveNet model outperforms the other two neural network models and the two naive baselines. 

\begin{table*}
  \caption{ The appliance-level mean absolute error (MAE) in unit of watt and signal aggregate error (SAE). Best results are shown in bold.}
  \vspace{6pt}
  \label{tab:result}
  \centering
  \begin{tabular}{c|c|cccccc}
\toprule
Metrics & Methods & Kettle & Microwave & Dishwasher & Washing M. & Overall \\
\midrule
MAE & Always-zero &10.157 &4.386 &20.784 & 6.189 & 10.378$\pm$6.359\\
& CNN \cite{zhang2018sequence} &5.454 &4.002 &21.014 &4.970 & 8.860$\pm$7.036 \\
& RNN  &4.839 &3.696 &15.261 &3.602 & 6.849$\pm$4.880\\
& WaveNet &\textbf{4.726} &\textbf{3.686} &\textbf{10.296} &\textbf{3.080} & \textbf{5.446$\pm$2.860} \\
\midrule
SAE  & Always-mean &1.347 &0.713 &1.121 & 2.121 & 1.325$\pm$0.512\\
& CNN \cite{zhang2018sequence} &0.258 &0.797 &0.976 &0.440 & 0.617$\pm$0.283\\
& RNN  &0.249 &\textbf{0.644} &0.377 &\textbf{0.208} & 0.369$\pm$0.170\\
& WaveNet &\textbf{0.224} &0.666 &\textbf{0.192} & 0.267 & \textbf{0.337$\pm$0.191}\\
\bottomrule
\end{tabular}
\end{table*}

Among the four appliances, \textit{microwave} is the only one that all the three neural network models achieve comparable results as that of the model of always-zero and always-mean. A closer inspection of the REFIT dataset shows that microwaves were mostly operated on either the off mode or the standby mode (0 to 5 watts) and the latter composes of more than 99.6\% of the readings which is the highest among the four appliances.   

\begin{figure}[!h]
  \centering
  \centerline{\includegraphics[width=0.99\columnwidth]{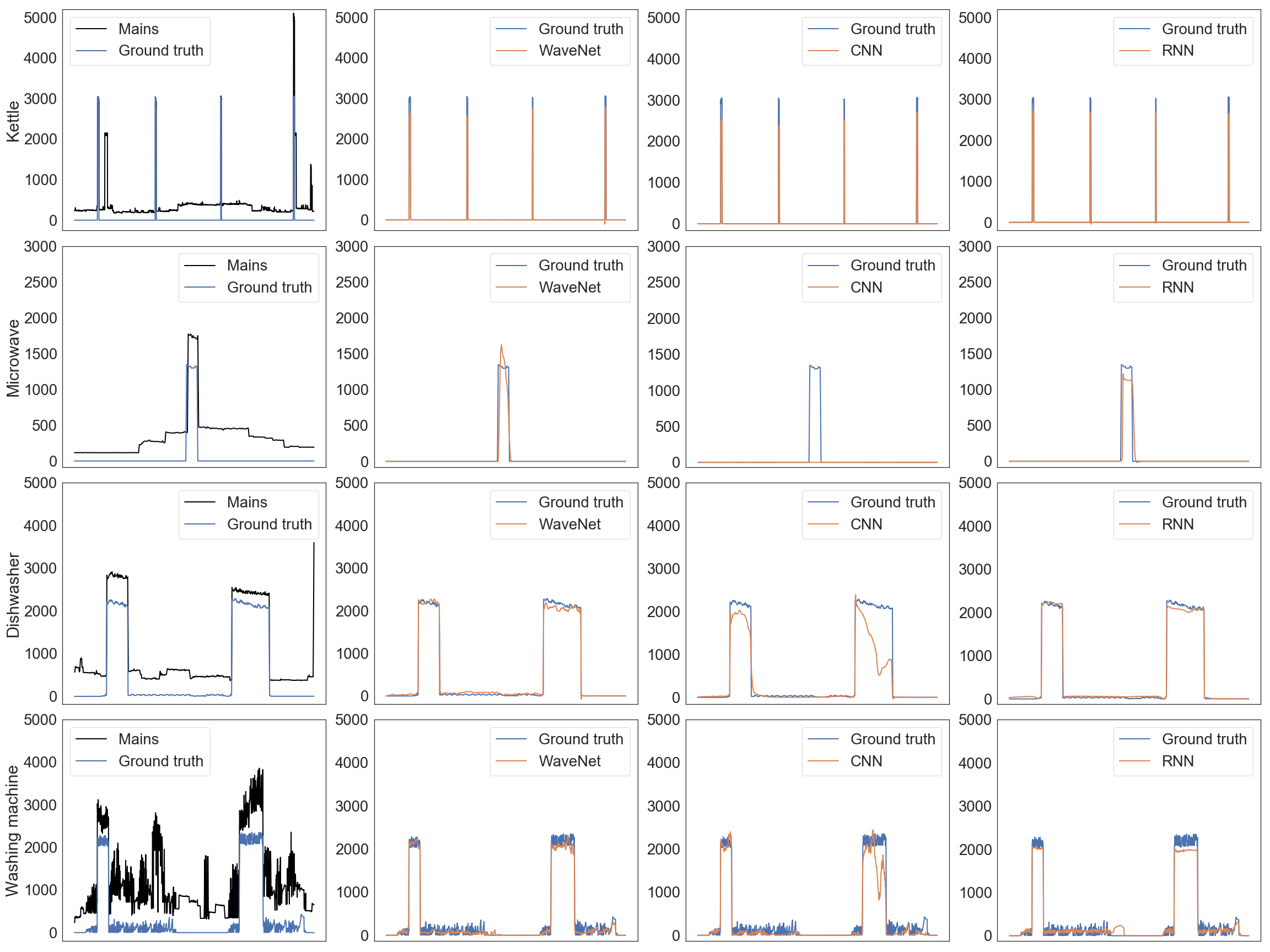}}
  \caption{Excerpts of disaggregation results with respect to the CNN, RNN and WaveNet model achieved the best MAE.}
  \label{fig:visualisation}
\end{figure}

To have a visual understanding of the disaggregation results, Figure \ref{fig:visualisation} shows for each type of appliance an excerpt of the predictions together with the target values with respect to the CNN, RNN and WaveNet models that achieve the best MAE (as shown in Table \ref{tab:result}). It can be seen that the disaggregation results for \textit{kettle} are similar among the three models. As for \textit{microwave}, the WaveNet model has some predictions that are larger than the target values while the predictions of the RNN model are mostly smaller than the target values. The CNN model however does not recognise the operation of the microwave, which is in line with the fact that the corresponding MAE of the CNN model is similar to that of the always-zero model. As for \textit{dishwasher} and \textit{washing machine}, the predictions of the WaveNet model are finer and closer to the target values compare to that of the RNN model, and the predictions of the CNN model are much noisier than the other two models.    

\begin{figure}
  \centering
  \centerline{\includegraphics[width=0.99\columnwidth]{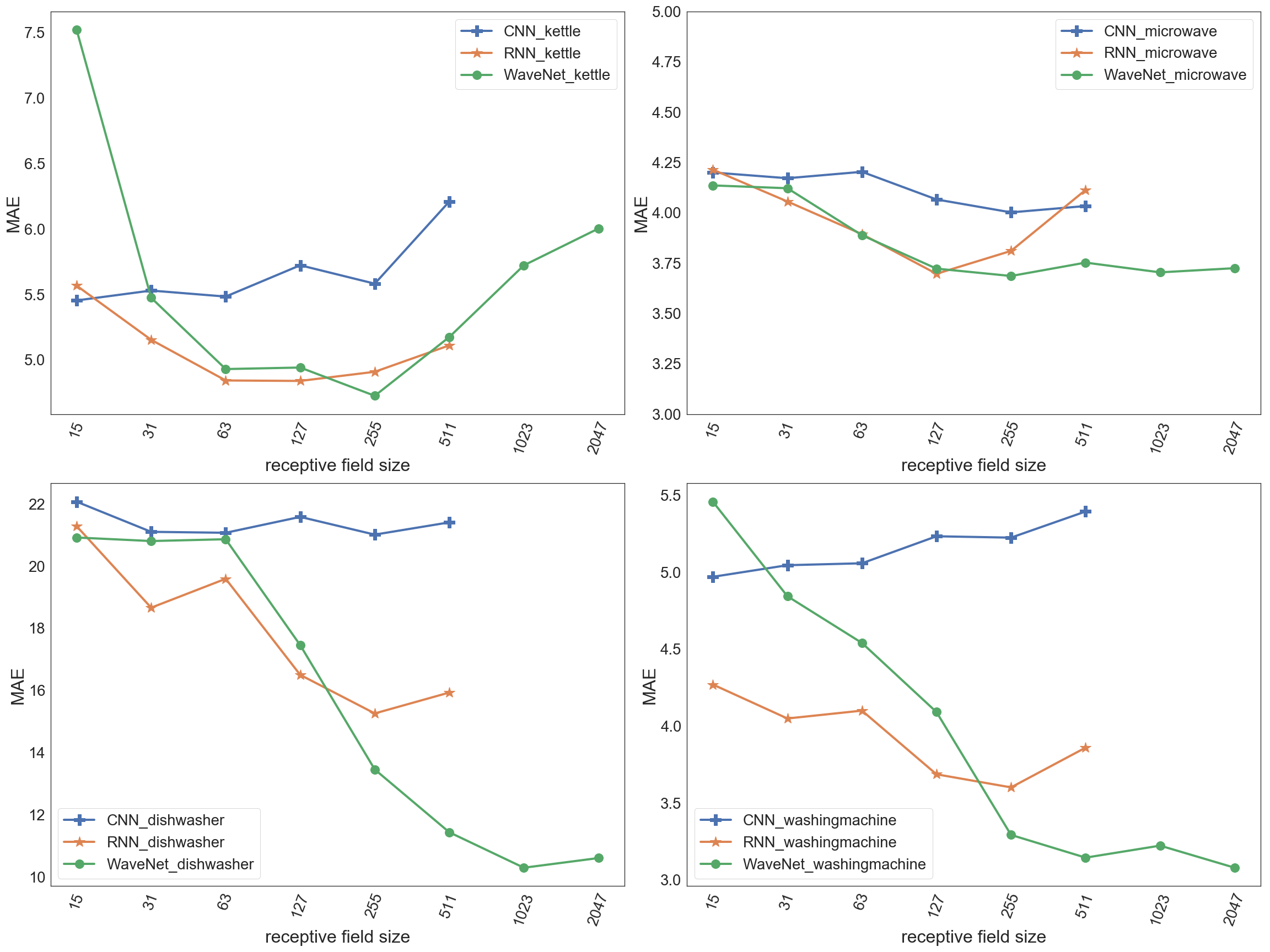}}
  \caption{Mean Absolute Error (MAE) of the CNN, RNN and WaveNet models with different receptive field sizes for the four appliances. }
  \label{fig:results-seqlen}
\end{figure}

To investigate how the length of input sequences influences the model performance, we compare the MAEs achieved by the CNN, RNN and WaveNet models as shown in Figure \ref{fig:results-seqlen}. Note that the receptive field size of \textit{1023} and \textit{2047} were only applied to the WaveNet models for the sake of computation cost. The size of the receptive field in general does not have much influence on the performance of the CNN models compared to the other two groups of models. The RNN models in general achieve better MAE when the size of the receptive field gets longer but when the receptive filed size is larger than 255 the performance gets worse. As for the WaveNet models, there is a clear tendency that its performance is getting better with longer receptive fields in the cases of \textit{dishwasher} and \textit{washing machine}. An explanation is that dishwashers and washing machines have relatively longer period of operation and the models need more information to capture the energy consumption patterns. In the case of \textit{kettle}, the WaveNet models achieve better MAE with the size of the receptive field getting longer up to 255 and thereafter the performance starts getting worse. This may be explained by the fact that kettles usually have a short operation time and any longer receptive field will introduce too much noise. 

\begin{figure}
  \centering
  \centerline{\includegraphics[width=0.5\columnwidth]{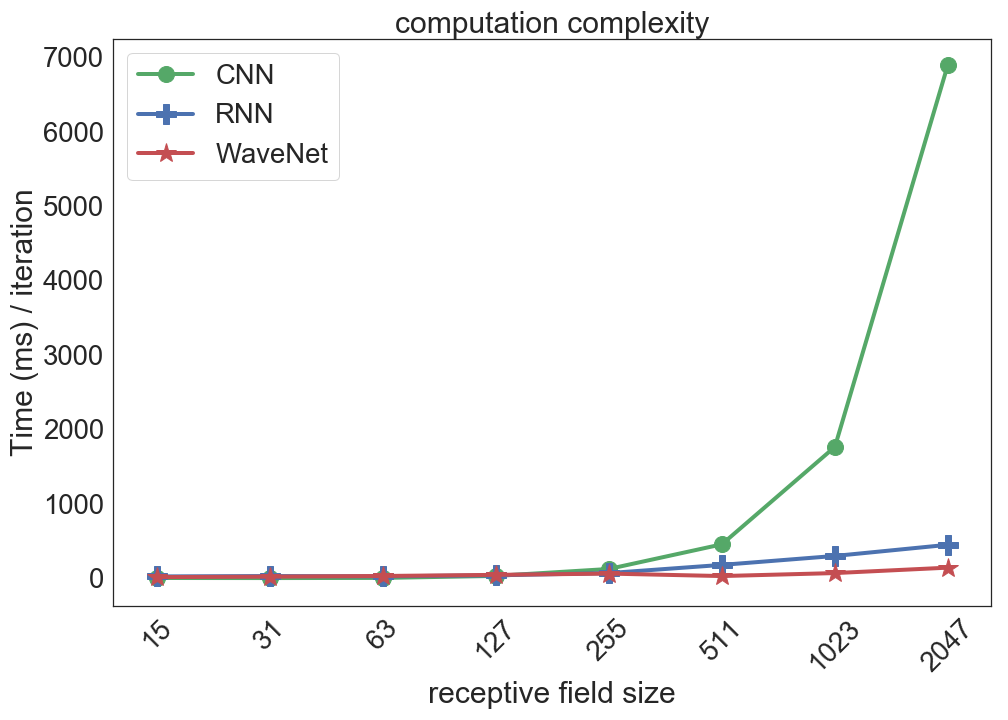}}
  \caption{Computation time per iteration for the CNN, RNN and WaveNet models with different receptive field sizes. }
  \label{fig:results-computation}
\end{figure}

Training efficiency is also an important factor when comparing models.
Figure \ref{fig:results-computation} shows the training time of the CNN, RNN and WaveNet models. We can see that when the receptive field size is above 511 the computation time of the CNN models increases quadratically. The WaveNet models have the lowest computation cost when the receptive field size becomes substantially large ($\geq$ 511) among the three groups of models. Furthermore, the WaveNet models converge much quicker than the other two groups of models. For example, for \textit{washing machine}, the number of iterations that the CNN models and the RNN models needed for training until convergence is more than 4 times of that needed by the WaveNet models. 

\begin{figure}[!h]
  \centering
  \centerline{\includegraphics[width=0.99\columnwidth]{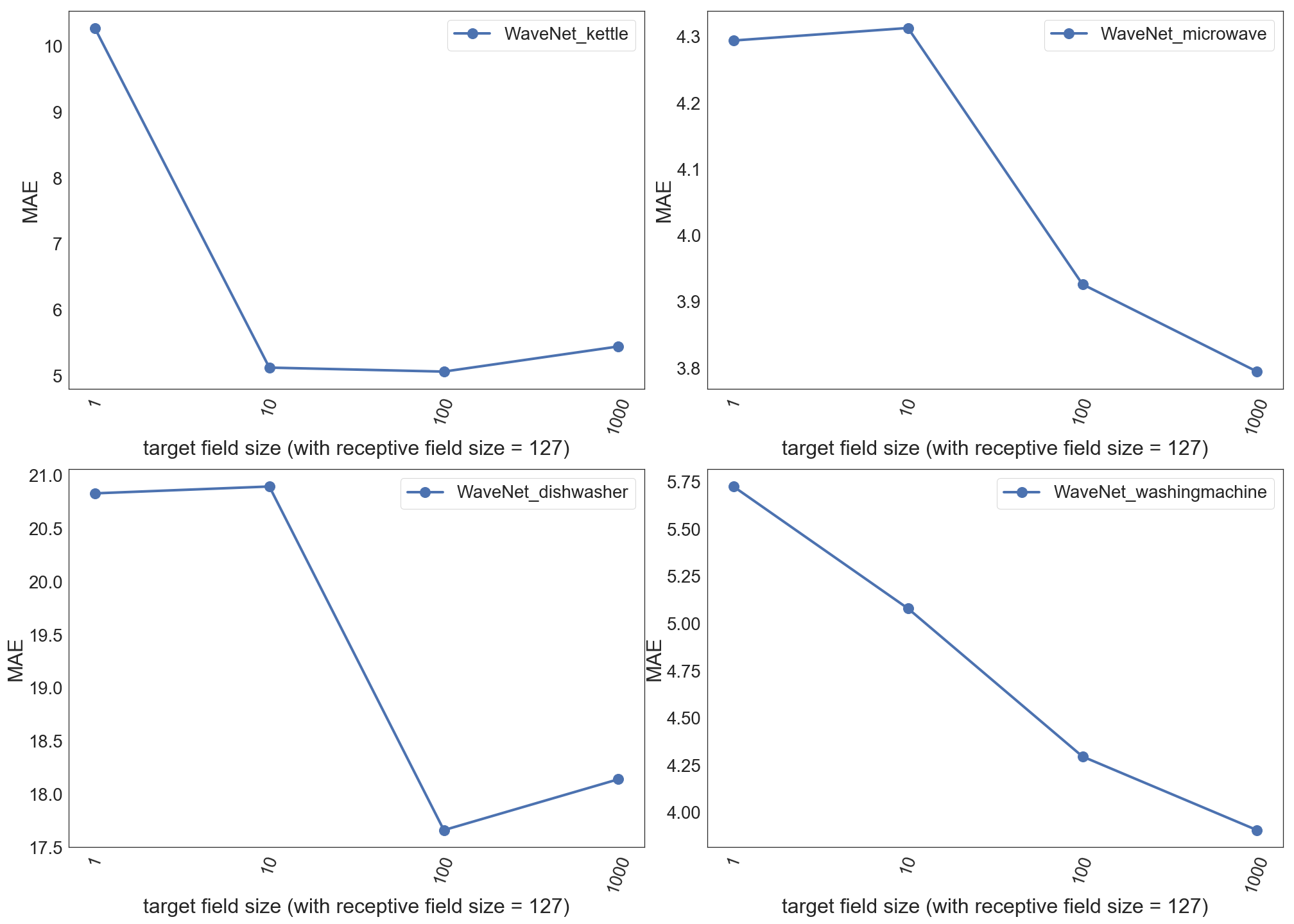}}
  \caption{Mean absolute error (MAE) of the WaveNet models with different target field sizes. }
  \label{fig:results-width}
\end{figure}

Target field size is a parameter that is worth exploring as well. In Figure \ref{fig:results-width} we show the relation between the target field size and the performance of the WaveNet models in terms of MAE with a fixed receptive field size of 127. We can see that for all the four appliances there is a tendency that the longer target fields achieve better MAE. This is because the longer target fields provide more training samples per mini-batch which is similar to the effect of applying a larger batch size and is more likely to converge to global optima. Comparing to using a larger batch size, the computation efficiency of using longer target fields is much higher due to shared computations.

\subsection{Experimental Results For On/off Detection}
\subsubsection{Experiment Setup}
As for the task of on/off detection, our aim is to compare the performance of the regression based learning framework and the classification based learning framework as proposed in Section \ref{sec:learning_onoff}. To this end, we trained two groups of WaveNet models following the two learning frameworks as it has been shown in the previous subsection that WaveNets achieve the best performance for the task of energy disaggregation. We experimented with a range of receptive field sizes \textit{15}, \textit{31}, \textit{63}, \textit{127}, \textit{255}, \textit{511}, \textit{1023}, \textit{2047}. The Adam optimizer is used with a learning rate of 0.001 to minimise the loss function shown in Equation \ref{eq:loss} for the regression based learning framework and in Equation \ref{binary_crossentropy} for the classification based learning framework. 

\subsubsection{Result Analysis}
\begin{figure}[!h]
  \centering
  \centerline{\includegraphics[width=0.99\columnwidth]{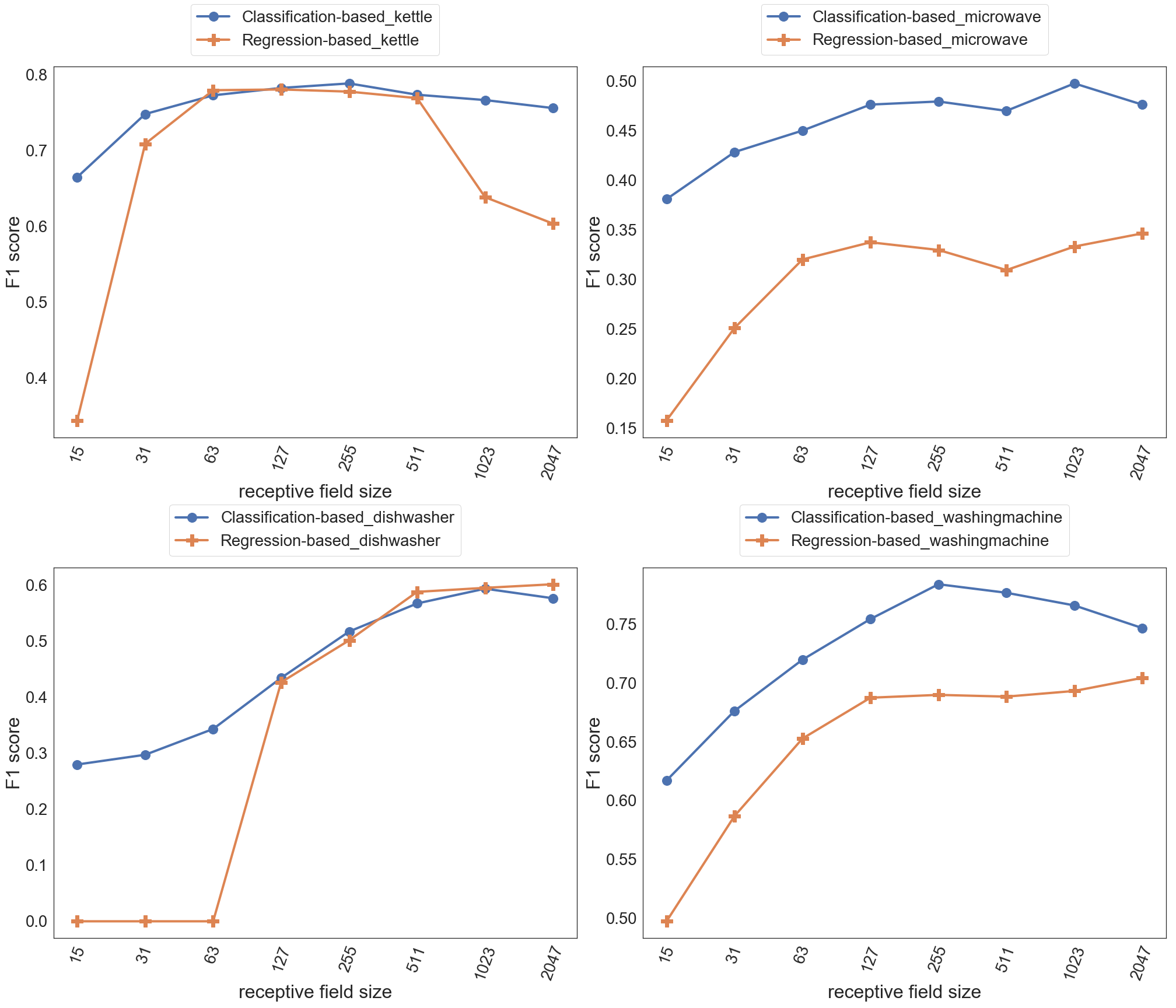}}
  \caption{F1 score of the WaveNet models trained following the two learning frameworks.}
  \label{fig:classifi-results}
\end{figure}

Figure \ref{fig:classifi-results} shows the F1 scores obtained by the WaveNet models trained respectively under the two learning frameworks with an increasing receptive field size. For the binary classifier under the classification based learning framework, we use a cut-off probability of 0.3, i.e., when the classifier outputs a value larger than 0.3 we consider the appliance is in the on state otherwise in the off state. 

We can see that in the case of \textit{kettle} and \textit{dishwasher} the classification based learning framework achieves better F1 score than the regression based learning framework when the receptive field size (or the number of dilated convolutional layers) is small. With larger receptive fields the two frameworks perform similarly. As for \textit{microwave} and \textit{washing machine}, the classification based learning framework achieves better F1 score than the regression based learning framework over all the receptive field sizes.

\section{Conclusions}\label{conclusion}
In this paper, we investigated the problem of energy disaggregation together with the problem of appliance on/off detection. Firstly, we formalised both problems and illustrated the learning/training paradigms used in the literature, which motivated us to introduce the fast-sequence-to-point learning paradigm. By comparing with CNN models and RNN models, we studied the application of the recently proposed WaveNet models to the problem of energy disaggregation. With an evaluation on a real-world dataset, we showed that our disaggregation models based on WaveNets outperform the previous works based on CNNs and RNNs. The empirical evidence demonstrates WaveNets' superiority in handling long sequences. By an extensive experiment with input sequences of varying receptive field sizes, we have shown how the receptive field size affects the disaggregation performance for different appliances. Furthermore, we studied the problem of appliance on/off detection as a natural continuation of the disaggregation problem and investigated the performance of two learning frameworks: (1) a regression based learning framework utilising the results from energy disaggregation and (2) a classification based learning framework that directly trains a binary classifier. We showed empirically that the classification based learning framework outperforms the regression based learning framework in terms of F1 score. This indicates that for applications targeting at appliance on/off states, directly training a binary classifier would be a better choice.

For future work, we intend to explore the use of prior knowledge to enhance the learning of WaveNet models. Another interesting direction for future work is to make use of the on/off states of the appliances to improve the results of energy disaggregation. For example, we could use the on/off states of an appliance to condition the predictions of the amount of energy the appliance consumes.  

\section{Acknowledgement}

This work was carried out as part of the ``HomeSense: digital sensors for social research'' project funded by the Economic and Social Research Council (grant ES/N011589/1) through the National Centre for Research Methods. Qiuqiang Kong was supported by EPSRC grant EP/N014111/1 ``Making Sense of Sounds'' and a Research Scholarship from the China Scholarship Council (CSC) No. 201406150082.

\bibliographystyle{ACM-Reference-Format}
\bibliography{bibliography}

\end{document}